# KReF: Training-Free Retrieval for Long-Term Time-Series Forecasting and Predictive Uncertainty

**Yang Zhang and Rui Su**

**Abstract**

Probabilistic long-term time-series forecasting commonly relies on trained models. Training-free conformal methods typically construct intervals around a pre-existing point forecaster and do not natively represent a complete predictive distribution; sequential variants additionally suffer from increasingly delayed feedback at long horizons. We propose KReF, a training-free retrieval framework that treats retrieved historical futures as a query-local empirical predictive distribution. After robust preprocessing, KReF embeds each lookback using handcrafted statistics or frozen random Fourier features and retrieves similar historical lookback-future pairs. Their similarity weights directly define predictive masses, quantiles, CRPS, and a weighted-mean point forecast. KReF further uses the observed query lookback to construct a probability-integral-transform map and applies validation-selected expansion and shrinkage rates to adapt interval boundaries. Across six LTSF benchmarks and four horizons, KReF obtains the lowest CRPS in all 12 dataset-embedding settings and the lowest IS90 in 9 settings. Without gradient-based fitting, its point forecasts also match or surpass trained baselines on two of six datasets. An archive-oracle analysis further reveals substantial headroom under finer horizon- and channel-wise routing. These results establish retrieval as a useful and underexplored inductive bias for LTSF.

## 1. Introduction

Probabilistic forecasting methods in time series forecasting are commonly implemented through trained parametric models or probabilistic prediction heads. Training-free conformal methods provide an alternative, but they typically calibrate intervals around a pre-existing point forecaster rather than natively representing a complete predictive distribution. Sequential methods such as ACI (Gibbs and Candès 2021) additionally rely on error feedback that becomes increasingly delayed in long-horizon multi-step forecasting. Meanwhile, most LTSF point forecasters also depend on iterative parameter optimization and task-specific loss functions. These limitations motivate a unified question: can we construct a training-free forecasting mechanism that independently produces both point and distributional predictions?

Our key observation is that historical futures already constitute candidate outcomes for the query future. Given historical lookback-future pairs, lookbacks similar to the current query identify a local collection of plausible future trajectories. Their weighted empirical distribution directly provides predictive probability masses, quantiles, and prediction intervals, while their weighted aggregation provides a point forecast. This view motivates KReF, a training-free retrieval framework for multivariate LTSF.

KReF robustly preprocesses each lookback, embeds it through two embedding schemes and retrieves the nearest historical lookback-future pairs. Similarity weights over the retrieved futures define the query-local empirical predictive distribution and its weighted-mean point forecast. Within the same uncertainty construction, KReF uses the observed query lookback to form a lookback-conditioned PIT probability map and validation-selected expansion and shrinkage rates to adapt interval boundaries. KReF therefore operates as a standalone point and probabilistic forecaster, but does not require its uncertainty representation to be tied to its internal KNN mean. Our contributions are as follows:

- We introduce a training-free retrieval framework in which weighted historical futures form a query-local empirical predictive distribution, enabling point forecasts, predictive quantiles, and direct CRPS evaluation.
- Across six benchmarks, KReF achieves the lowest CRPS in all 12 dataset-embedding settings and the lowest IS90 in 9 settings, while its training-free point forecasts surpass trained baselines on two datasets. Analysis further exposes the difficulty of delayed online calibration in LTSF, while an archive-oracle study reveals substantial headroom for finer-grained retrieval.

## 2. Related Work

**Retrieval, analog forecasting, and random features.** Analog forecasting retrieves historical states and uses their observed successors as forecasts; analog ensembles extend this principle to probabilistic weather prediction (Delle Monache et al. 2013). Recent retrieval-based time-series methods instead pass retrieved references to trained synthesis or diffusion models (J. Liu et al. 2024). Similarity is computed from handcrafted summaries or frozen Random Fourier Features, which approximate shift-invariant kernels (Rahimi and Recht 2007).

**Uncertainty Estimation and Calibration.** Conformal prediction provides marginally valid intervals under exchangeability (Vovk et al. 2005; Lei et al. 2018), while sequential extensions adapt to temporal shift through online miscoverage control (Xu and Xie 2021; Gibbs and Candès 2021). C-PID, ECI and COP refine this controller view (Angelopoulos et al. 2023; Wu et al. 2025; Hu et al. 2026), and Hallberg Szabadváry (2024) extends ACI to multi-step forecasts using lead-specific states. These methods calibrate intervals around a given point predictor and receive increasingly delayed feedback at longer leads.

Nearest-neighbor uncertainty methods use KNN either as the underlying conformal predictor or to localize calibration samples (Papadopoulos et al. 2011; Guan 2023; Tajmouati et al. 2024), while a recent preprint further studies online localized calibration for heterogeneous sequential data (Lai and Raskutti 2026). These methods calibrate residuals or fitted distributions around a KNN point forecast. KReF instead treats retrieved multi-step futures as the query-local predictive distribution itself and reweights its probability masses using the observed query lookback. Predictive-distribution recalibration has likewise composed a base CDF with a PIT probability map (Kuleshov et al. 2018), with Cal-PIT learning covariate-dependent maps for instance-wise calibration (Dey et al. 2025). KReF adopts this probability-map principle but constructs its query-specific map without fitting a calibrator or observing the query future.

**Long-term time-series forecasting.** LTSF models span patch- and variate-based Transformers such as PatchTST (Nie et al. 2023) and iTransformer (Y. Liu et al. 2024), multiscale mixers such as TimeMixer++ (S. Wang et al. 2025), and efficient frequency or linear architectures such as DLinear (Zeng et al. 2023), FilterTS (Y. Wang et al. 2025) and MixLinear (Ma et al. 2026). Recent methods exploit additional structure: CometNet guides learned experts with contextual motifs (Wang et al. 2026), whereas CGTFra combines frequency resampling, dynamic graphs, and information-theoretic alignment for inter-series dependency modeling (Yu et al. 2026). All remain trained parametric forecasters; KReF directly reuses retrieved historical futures without parameter optimization.

# 3. Method

## Task Formulation

Let $Z_{1:T} \in R^{T\times C}$ denote a multivariate time series with $\boldsymbol{C}$ variables. Given a lookback length $\boldsymbol{L}$ and a prediction horizon $\boldsymbol{H}$, each forecasting instance consists of a lookback window $X_i \in R^{L\times C}$ and its corresponding future trajectory $Y_i \in R^{H\times C}$. The goal of long-term time-series forecasting is to predict $Y_i$ from $X_i$.

We additionally consider probabilistic forecasting. For each horizon-channel coordinate, the target is a conditional predictive distribution $F_{i,h,c}(v) = Pr\big(Y_{i,h,c} \leq v \mid X_i\big)$.

from which a point forecast, predictive quantiles, and a nominal-$\alpha$ prediction interval $I_{i,h,c}^{(\alpha)} = \big[l_{i,h,c}^{(\alpha)}, u_{i,h,c}^{(\alpha)}\big]$ can be derived. We seek empirical coordinate-wise coverage close to $\alpha$ : $\frac{1}{|\mathcal{E}|}\sum_{(i,h,c)\in\mathcal{E}} I\left\{Y_{i,h,c} \in I_{i,h,c}^{(\alpha)}\right\} \approx \alpha$, where $\mathcal{E}$ denotes valid test sample-horizon-channel coordinates.

## Robust Window Preprocessing

Before constructing lookback embeddings, we apply a robust preprocessing procedure to reduce the effect of extreme values and unstable per-sample normalization. This step has two components: quantile clipping and safe normalization.

First, for each channel, we estimate lower and upper clipping thresholds from the training split as empirical $q$ and ($1$-$q$) quantiles. The lookback and future values used as retrieval database elements are clipped to this range.

Second, each clipped lookback window is normalized per sample and per channel. Let $\mu_i \in R^{1\times C}$ and $\sigma_i \in R^{1\times C}$ denote the channel-wise mean and standard deviation of the clipped lookback $X_i$. Standard per-sample z-score normalization can be unstable when a channel is nearly constant: a tiny standard deviation may amplify small deviations into very large normalized values. To avoid this, we use a safe scale $\bar{\sigma}_i$. For each sample and channel,

$$\bar{\sigma}_{i,c} = \begin{cases} \sigma_{i,c}, if\ \sigma_{i,c} > \epsilon \\ 1, otherwise \end{cases}$$

where $\epsilon$ is a small threshold. The normalized lookback is then $\tilde{x}_i = (X_i - \mu_i)/\bar{\sigma}_i$.

The corresponding future trajectory is represented in the same normalized coordinate system, $\tilde{Y}_i = (Y_i - \mu_i)/\bar{\sigma}_{i,c}$

At prediction time, after obtaining a normalized forecast($\hat{\tilde{Y}}_i$), we map it back to the original scale by $\hat{Y}_i = \hat{\tilde{Y}}_i \odot \bar{\sigma}_i + \mu_i$.

## Lookback Embedding

The embedding function $\phi(\cdot)$ maps a robustly normalized lookback window $\tilde{x}_i$ to a vector representation used for nearest-neighbor retrieval. We study two training-free embeddings: handcrafted statistical embeddings and Random Fourier Feature embeddings.

***Handcrafted Statistical Embedding***

The handcrafted embedding summarizes each channel of the normalized lookback using a fixed set of descriptive statistics. For each channel, we construct an 11-dimensional summary vector consisting of: the last value $\tilde{x}_{L,c}$; means over the last 24 and 96 steps; the 96-step standard deviation and interquartile range; least-squares slopes over the last 24 and 96 steps; the last 96-step z-score; the difference between the last value and the 24-step mean; the correlation between the most recent 24-step segment and its preceding 24-step segment; and the value at lag 24. The slope over a trailing window of length $w$ is

$$b_{w,c} = \frac{\sum_{r=1}^{w}(r-\bar{r})\left(\tilde{x}_{L-w+r,c} - \overline{\tilde{x}_{w,c}}\right)}{\sum_{r=1}^{w}(r-\bar{r})^2}$$

For each channel $c$, let $s_1, \dots, s_M$ denote the fixed summary statistics. The handcrafted embedding is obtained by concatenating these statistics across all channels:

$$\phi_{\text{stat}}(X_i) = \text{concat}_{c=1}^{C}\left[s_1\left(\tilde{X}_{i,:,c}\right), \dots, s_M\left(\tilde{X}_{i,:,c}\right)\right]$$

This embedding's purpose is to expose interpretable similarity dimensions such as local level, volatility, trend, and short-term deviation.

***Random Fourier Feature Embedding***

Handcrafted statistics are interpretable, but they compress the lookback window into a small set of summaries and may discard fine-grained temporal structure. To preserve a richer nonlinear notion of similarity over the full lookback trajectory, we also use Random Fourier Features (RFF).

Let $z_i = \text{vec}\left(\tilde{X}_i\right) \in R^{LC}$ be the flattened normalized lookback. We sample a frozen random matrix $W \in R^{LC \times D}$ and phase vector $b \in R^{D}$, where

$$W_{jk} \sim N(0, \sigma_{\text{rff}}^{-2}), \quad b_k \sim \text{Uniform}(0, 2\pi)$$

The RFF embedding is

$$\phi_{\text{rff}}(X_i) = \sqrt{2/D}\cos(z_i W + b)$$

By the random feature approximation of shift-invariant kernels, inner products in this feature space approximate Gaussian-kernel similarity between normalized lookback trajectories. In our method, $W$ and $b$ are sampled once and kept fixed. The bandwidth $\sigma_{\text{rff}}$ is selected using the median-distance heuristic over historical lookbacks.

Each embedding can be used independently; we also evaluate a lightweight stacking variant that combines their predictions using validation data.

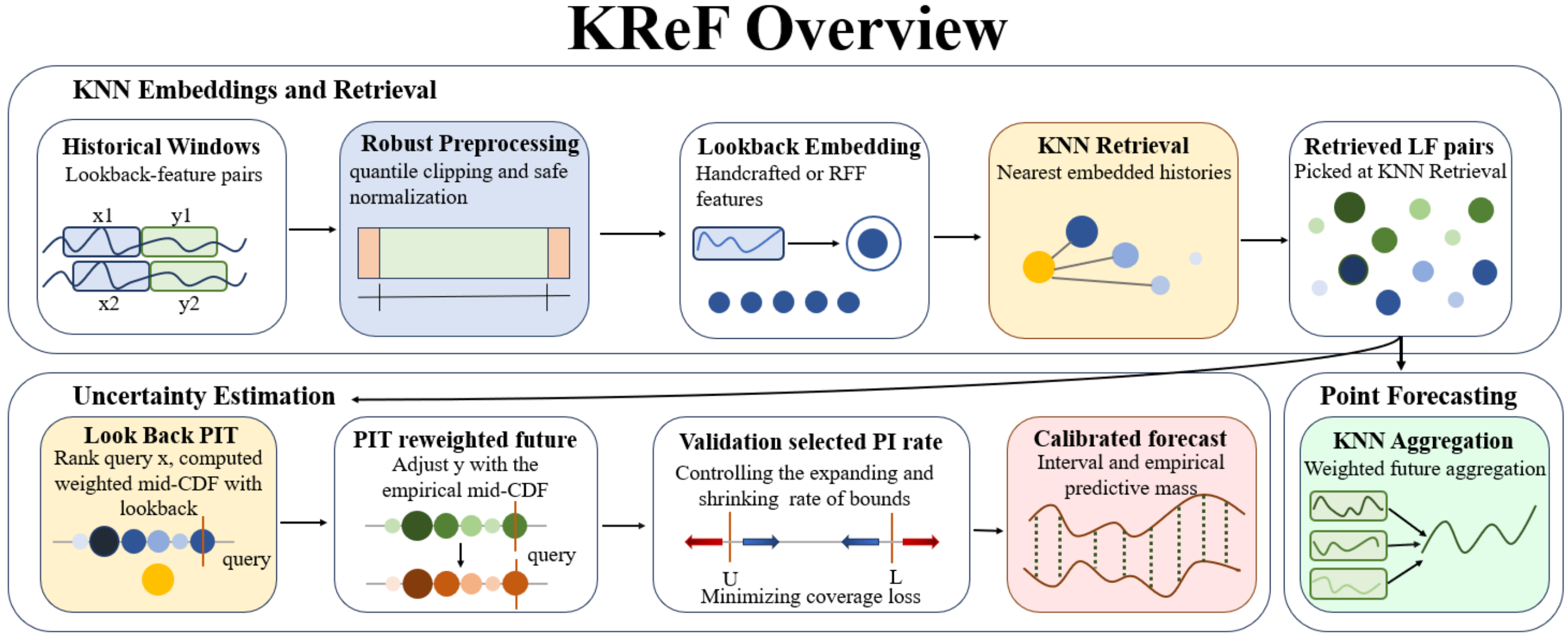

Figure 1: Overview of KReF. Historical lookback-future pairs are embedded and retrieved to support point forecasting and uncertainty estimation.

## KNN Forecasting

Given an embedding function $\phi(\cdot)$, we construct a retrieval database from historical lookback-future pairs. For each database instance $j$, we store its lookback embedding $\phi(X_j)$ and its normalized future trajectory $\tilde{Y}_j$. For a query lookback $X_i$, we compute its embedding $\phi(X_i)$ and retrieve the $K$ database instances with the largest cosine similarity,

$$\text{sim}(i,j) = \frac{\phi(X_i)^{\top}\phi(X_j)}{|\phi(X_i)|_2 |\phi(X_j)|_2}$$

Let $N_K(i)$ denote the set of retrieved neighbors. We convert similarities into softmax weights,

$$w_{ij} = \frac{\exp(\text{sim}(i,j)/\tau)}{\sum_{m \in N_K(i)} \exp(\text{sim}(i,m)/\tau)}, \quad j \in N_K(i)$$

where $\tau$ is a temperature parameter. Smaller $\tau$ produces sharper weights and makes the prediction rely more heavily on the nearest neighbors; larger $\tau$ averages over a broader local neighborhood.

The normalized forecast is the weighted average of the retrieved future trajectories: $\hat{\tilde{Y}}_i = \sum w_{ij} \tilde{Y}_j$

Finally, the prediction is mapped back to the original scale using the query lookback's normalization statistics:

$$\hat{Y}_i = \hat{\tilde{Y}}_i \odot \bar{\sigma}_i + \mu_i$$

The hyperparameters used, such as $K$, $\tau$, and the clipping quantile, can be selected on the validation split.

Because the method contains no iterative optimization tied to a particular lookback length, it can also be extended to multiple lookback lengths without retraining. Given a set of lookback lengths $\mathcal{L}$, we obtain one retrieval forecast $\hat{Y}_i^{(L)}$ for each $L \in \mathcal{L}$, and combine them by validation-weighted averaging,

$$\hat{Y}_i = \sum_{L\in\mathcal{L}} w_L \hat{Y}_i^{(L)}, w_L = \frac{\exp\left(-\gamma_{\mathcal{L}} E_L^{\text{val}}\right)}{\sum_{L'\in\mathcal{L}} \exp\left(-\gamma_{\mathcal{L}} E_{L'}^{\text{val}}\right)},$$

where $E_L^{\text{val}}$ is the validation error of lookback length $L$.

In the offline setting, the retrieval database contains historical training and validation instances. In the online setting, the database is updated over time with previously observed test instances. We use a pessimistic online protocol: a test instance is added to the database only after its full future horizon becomes observable. This avoids leaking unavailable future values while allowing the retrieval database to adapt to temporal distribution shift.

## Uncertainty Estimation

The retrieved futures naturally define a query-local empirical predictive distribution. For query i, let $N_K(i)$ denote its retrieved neighbors and $w_{ij}$ the similarity weights defined above. At horizon-channel coordinate $(h,c)$, the base empirical CDF in the normalized space is

$$F_{i,h,c}^{\text{B}}(z) = \sum_{j\in N_K(i)} w_{ij}\ \mathrm{I}\left(\tilde{Y}_{j,h,c} \le z\right)$$

For nominal coverage $\alpha$, let $p_\alpha = \frac{1-\alpha}{2}$, The corresponding base interval is:

$$L_{i,h,c}^{\text{B}}(\alpha) = \left(F_{i,h,c}^{\text{B}}\right)^{-1}(p_\alpha), U_{i,h,c}^{\text{B}}(\alpha) = \left(F_{i,h,c}^{\text{B}}\right)^{-1}(1-p_\alpha)$$

These weighted quantiles describe the local variation among the retrieved futures, but their probability masses may be miscalibrated when the query occupies a systematically different position from its neighbors. We therefore use the fully observed query lookback to construct a query-specific probability-integral-transform adjustment.

***Lookback-conditioned PIT reweighting.***

For every valid lookback coordinate $(l, c)$, we compute the weighted midrank of the query value among the retrieved lookbacks: $R_{i,l,c} = \sum_{j\in N_K(i)} w_{ij}\left[\mathrm{I}\left(\tilde{X}_{j,l,c} < \tilde{X}_{i,l,c}\right) + \frac{1}{2}\mathrm{I}\left(\tilde{X}_{j,l,c} = \tilde{X}_{i,l,c}\right)\right]$

Pooling these ranks over all valid lookback-channel coordinates gives the empirical mid-CDF.

$$G_i(u) = \frac{1}{|\mathcal{V}_i|}\sum_{(l,c)\in\mathcal{V}_i}\left[\mathrm{I}\left(R_{i,l,c} < u\right) + \frac{1}{2}\mathrm{I}\left(R_{i,l,c} = u\right)\right]$$

where $\mathcal{V}_i$ is the set of valid lookback-channel coordinates. Intuitively, $G_i$ summarizes where the current lookback lies within its retrieved neighborhood. We transfer this rank information to every future coordinate by composing the base future CDF with $G_i$:$F_{i,h,c}^{\text{LB}}(z) = G_i\left(F_{i,h,c}^{\text{B}}(z)\right)$.

For a discrete implementation, sort the retrieved future values at $(h,c)$ and define their cumulative base masses as $C_{i,k,h,c} = \sum_{r=1}^{k} w_{i(r)}, C_{i,0,h,c} = 0$.

The PIT-adjusted mass assigned to the $k$-th sorted future is then $\tilde{w}_{i(k),h,c} = G_i\left(C_{i,k,h,c}\right) - G_i\left(C_{i,k-1,h,c}\right)$

Thus, the support remains the retrieved historical futures, while their probability masses become conditional on the position of the current lookback.

The adjusted masses therefore provide predictive quantiles and a native empirical CRPS: $CRPS\left(F_{i,h,c}^{\text{LB}}, \tilde{Y}_{i,h,c}\right) = \sum_j \tilde{w}_{ij,h,c}\left|\tilde{Y}_{j,h,c} - \tilde{Y}_{i,h,c}\right| - \frac{1}{2}\sum_{j,k}\tilde{w}_{ij,h,c}\tilde{w}_{ik,h,c}\left|\tilde{Y}_{i,h,c} - \tilde{Y}_{k,h,c}\right|$

The KNN point forecast remains the similarity-weighted mean defined previously; PIT reweighting is used only for uncertainty estimation.

***Validation-selected interval adjustment.***

Let $L^{LB}(\alpha)$ and $U^{LB}(\alpha)$ denote the lower and upper quantiles of $F^{LB}(\alpha)$. Although Lookback-PIT can move the interval in a query-dependent direction, its displacement may be insufficient under strong distribution shift or unnecessarily aggressive when the base interval is already conservative. We therefore introduce an expansion rate $\eta_e \ge 1$ and a shrinkage rate $0 \le \eta_s \le 1$.

Define $\Delta_{i,h,c}^{-} = L_{i,h,c}^{\text{LB}} - L_{i,h,c}^{\text{B}}$, $\Delta_{i,h,c}^{+} = U_{i,h,c}^{\text{LB}} - U_{i,h,c}^{\text{B}}$, The final normalized interval is:

$$L_{i,h,c} = L_{i,h,c}^{B} + \begin{cases}\eta_e\Delta_{i,h,c}^{-}, & \Delta_{i,h,c}^{-} < 0\\ \eta_s\Delta_{i,h,c}^{-}, & \Delta_{i,h,c}^{-} \ge 0\end{cases}$$

$$U_{i,h,c} = U_{i,h,c}^{B} + \begin{cases}\eta_s\Delta_{i,h,c}^{+}, & \Delta_{i,h,c}^{+} < 0\\ \eta_e\Delta_{i,h,c}^{+}, & \Delta_{i,h,c}^{+} \ge 0\end{cases}$$

Thus, outward movements are amplified by $\eta_e$, whereas inward movements are moderated by $\eta_S$. Large expansion rates can occasionally produce extreme intervals when the base weighted-quantile interval is narrow. We therefore define an optional trust-region safeguard. Let $b_{i,h,c}^{\text{B}} = \frac{U_{i,h,c}^{\text{B}} - L_{i,h,c}^{\text{B}}}{2}$, be the base half-width. For a clipping factor $\kappa$, the reported bounds are

$$L_{i,h,c} = \max\left\{L_{i,h,c}, L_{i,h,c}^{\text{B}} - (\kappa-1)b_{i,h,c}^{\text{B}}\right\}$$
$$U_{i,h,c} = \min\left\{U_{i,h,c}, U_{i,h,c}^{\text{B}} + (\kappa-1)b_{i,h,c}^{\text{B}}\right\}$$

For every coverage level $\alpha$, we select $(\eta_e, \eta_s)$ on the validation split. Validation origins are forecast using a frozen train-only retrieval archive. For each candidate pair, let $\hat{e}_\alpha^-$ and $\hat{e}_\alpha^+$ be the empirical lower- and upper-tail miss rates. We minimize $\mathcal{J}_\alpha = |\hat{e}_\alpha^+ - p_\alpha| + |\hat{e}_\alpha^- - p_\alpha|$. Ties are resolved by smaller average width.

Finally, all interval boundaries are mapped back to the original scale using the query lookback's normalization statistics. Although KReF's uncertainty estimator does not require a point forecast, its predictive distribution can be paired with either an external point forecaster or KReF's own weighted mean or median.

## 4. Experiments

### Experimental Setup

**Datasets**. We evaluate on six LTSF benchmarks: the four ETT datasets (ETTh1, ETTh2, ETTm1, ETTm2) from the ETT collection (Zhou et al. 2021), Weather and Exchange (Exch.) in the Autoformer benchmark collection (Wu et al. 2021). The Exchange dataset was originally introduced for multivariate forecasting by Lai et al. (2018). We follow the standard chronological splits used in the LTSF literature: 6:2:2 for the ETT datasets and 7:1:2 for Weather and Exchange. ECL and Traffic are excluded because their dimensionality makes H=720 retrieval exceed our 24 GB memory budget.

**Forecasting settings**. We use a fixed lookback length L=96 and prediction horizons $H \in \{96,192,336,720\}$ . All channels are predicted jointly. All metrics are reported in globally standardized data space, following the standard LTSF evaluation convention.

**Method instantiation**. We evaluate the two training-free embeddings introduced in the Method section, together with their lightweight validation-based stacking. The RFF embedding uses an output dimension of 512.

**Hyperparameters.** The Exchange-rate dataset uses an empirical q-quantile of 0; the other datasets use an empirical q-quantile of 0.01. The original unclipped future values are kept for evaluation and prediction interval calibration. K and $\tau$ are selected on the validation split by minimizing validation MSE. We grid-search the number of neighbors $K \in \{20, 50, 100, 200, 500, 1000\}$ and the softmax temperature $\tau \in \{0.05, 0.1, 0.3, 0.5, 1, 5\}$ and report the online pessimistic variant. The interval layer selects an expansion rate $\eta_e \in \{1,1.25,1.5,2,3\}$ and a shrinkage rate $\eta_e \in \{0,0.25,0.5,0.75,1\}$, no clipping safeguard is used in the main experiments. Selection uses a frozen train-only archive, and all selected settings remain fixed during testing.

**Baselines**. For uncertainty estimation, we compare only with training-free methods that require neither a loss function nor iterative optimization. All methods use the same online-pessimistic KNN point forecast, so differences reflect only their uncertainty construction. Interval evaluation baselines include a static SplitConf (SpC; Lei et al. 2018), a causally expanding online variant (OL-SpC; cf. Tajmouati et al. 2024), and asynchronous multi-step adaptations of ACI, scorecaster-free Conformal PID, and COP (MS-ACI, MS-C-PI, and MS-COP; Hallberg Szabadváry 2024; Angelopoulos et al. 2023; Hu et al. 2026).

CRPS baselines include horizon-channel Gaussian residuals, static empirical residuals, and causally expanding empirical residuals (Gauss, ER-S, and ER-E), following the residual-bootstrap and adaptive residual-selection literature (Hyndman and Athanasopoulos 2018; Wang et al. 2022). Detailed constructions and hyperparameters are provided in Appendix A.

Point Forecasting baselines include five published LTSF models that span representative architectural designs: a patch-based Transformer (PatchTST, Nie et al. 2023), a variate-tokenized Transformer (iTransformer, Y. Liu et al. 2024), a multiscale neural mixer (TimeMixer++, S. Wang et al. 2025), a frequency-filtering linear model (FilterTS, Y. Wang et al. 2025) and a consistency-driven inter-variate dependency model (CGTFra, Yu et al. 2026). For the trained point-forecasting baselines, we report the results published in their original papers under the standard LTSF protocol.

**Metrics.** Point forecasts are evaluated with MSE and MAE; prediction intervals by coverage, width, and interval score; and predictive distributions by CRPS. Definitions of these metrics are provided in Appendix B.

**Implementation.** Retrieval embeddings, similarity search, top-K selection, neighbor aggregation, and KReF uncertainty construction are executed on a single NVIDIA GeForce RTX 3090 GPU. Data loading and serialization are performed on CPU.

### Experimental Results

**Uncertainty Estimation Results**. Table 1 reports four-horizon-average interval and distributional results. KReF obtains the lowest IS90 under both embeddings on ETTh1, ETTh2, ETTm1, and Exchange, and additionally on ETTm2 with the handcrafted embedding. OL-SpC achieves the lowest IS90 in the remaining three settings: ETTm2-RFF and both Weather embeddings.

The controller-based baselines generally approach nominal coverage but often require substantially wider intervals, especially on ETTm1 and Weather. Their horizon-specific states receive feedback only after the corresponding targets become observable, which can produce stale reactions and severe overexpansion at long leads. Although KReF does not uniformly attain nominal coverage, particularly on ETTm1 and Weather, it still improves coverage over SpC and remains much narrower than the controller-based methods.

The four-horizon average also conceals the most severe delayed-feedback case. At H=720 on Exchange, all methods under-cover, but KReF retains the highest Cov90 at 0.845 and 0.783 under handcrafted and RFF embeddings, compared with 0.742 and 0.681 for the strongest adaptive baseline (Appendix C1). Average coverage results at the 50% and 80% nominal levels are reported in Appendix C2.

For full-distribution forecasting, KReF achieves the lowest CRPS in all 12 dataset-embedding settings, reducing CRPS by 3.8%–11.3% relative to the strongest distributional baseline. ER-E improves upon ER-S in all 12 settings, confirming that causally incorporating newly observed residuals is useful. KReF nevertheless remains consistently stronger, suggesting that its query-specific retrieved distribution captures information unavailable to residual pools conditioned only on forecast coordinates and observation time.

| Dataset | Metric | Handcrafted | | | | | | RFF | | | | | |
|---|---|---|---|---|---|---|---|---|---|---|---|---|---|
| | | Ours | SpC | MS-ACI | MS-C-PI | MS-COP | SpC-OL | Ours | SpC | MS-ACI | MS-C-PI | MS-COP | SpC-OL |
| ETTh1 | Cov90 | 0.887 | 0.881 | 0.900 | 0.889 | 0.892 | 0.896 | 0.887 | 0.886 | 0.902 | 0.886 | 0.889 | 0.902 |
| | Width90 | 2.699 | 2.680 | 4.773 | 16.073 | 14.252 | 2.791 | 2.298 | 2.497 | 4.207 | 13.350 | 11.930 | 2.568 |
| | IS90 | **3.432** | 3.739 | 5.659 | 17.228 | 15.371 | 3.712 | **2.978** | 3.360 | 4.975 | 14.401 | 12.948 | 3.318 |
| ETTh2 | Cov90 | 0.905 | 0.908 | 0.901 | 0.892 | 0.893 | 0.913 | 0.907 | 0.911 | 0.901 | 0.891 | 0.893 | 0.914 |
| | Width90 | 1.961 | 2.111 | 4.209 | 9.749 | 9.036 | 2.095 | 1.855 | 2.089 | 4.024 | 9.048 | 8.407 | 2.061 |
| | IS90 | **2.575** | 2.674 | 4.827 | 10.518 | 9.781 | 2.639 | **2.455** | 2.639 | 4.628 | 9.794 | 9.127 | 2.601 |
| ETTm1 | Cov90 | 0.884 | 0.906 | 0.900 | 0.918 | 0.918 | 0.908 | 0.868 | 0.906 | 0.900 | 0.919 | 0.919 | 0.908 |
| | Width90 | 2.224 | 2.545 | 32.359 | 50.434 | 48.344 | 2.532 | 1.939 | 2.489 | 31.775 | 51.873 | 48.651 | 2.466 |
| | IS90 | **3.020** | 3.446 | 33.275 | 51.172 | 49.072 | 3.428 | **2.791** | 3.374 | 32.688 | 52.583 | 49.353 | 3.351 |
| ETTm2 | Cov90 | 0.902 | 0.899 | 0.900 | 0.916 | 0.917 | 0.908 | 0.902 | 0.912 | 0.900 | 0.916 | 0.917 | 0.914 |
| | Width90 | 1.667 | 1.896 | 10.713 | 22.324 | 21.765 | 1.863 | 1.968 | 2.023 | 10.131 | 21.523 | 21.163 | 1.964 |
| | IS90 | **2.196** | 2.456 | 11.221 | 22.818 | 22.256 | 2.373 | 2.472 | 2.496 | 10.636 | 22.039 | 21.670 | **2.426** |
| Weather | Cov90 | 0.886 | 0.845 | 0.898 | 0.908 | 0.908 | 0.871 | 0.873 | 0.844 | 0.898 | 0.909 | 0.910 | 0.872 |
| | Width90 | 2.113 | 1.684 | 10.539 | 18.530 | 17.975 | 1.765 | 2.044 | 1.702 | 10.766 | 20.376 | 19.260 | 1.777 |
| | IS90 | 2.738 | 2.557 | 11.157 | 19.105 | 18.548 | **2.518** | 2.737 | 2.603 | 11.399 | 20.952 | 19.844 | **2.544** |
| Exch. | Cov90 | 0.851 | 0.761 | 0.843 | 0.843 | 0.845 | 0.811 | 0.824 | 0.739 | 0.829 | 0.832 | 0.834 | 0.797 |
| | Width90 | 2.619 | 1.790 | 3.459 | 7.620 | 7.319 | 2.192 | 2.463 | 1.663 | 3.347 | 7.428 | 7.151 | 2.089 |
| | IS90 | **3.496** | 3.745 | 4.787 | 9.197 | 8.876 | 3.760 | **3.470** | 4.127 | 5.016 | 9.187 | 8.899 | 3.982 |
| Dataset | CRPS | Ours | Gauss | | ER-S | | ER-E | Ours | Gauss | | ER-S | | ER-E |
| ETTh1 | CRPS | **0.436** | 0.466 | | 0.456 | | 0.453 | **0.379** | 0.411 | | 0.402 | | 0.398 |
| ETTh2 | CRPS | **0.282** | 0.305 | | 0.309 | | 0.297 | **0.271** | 0.294 | | 0.299 | | 0.288 |
| ETTm1 | CRPS | **0.351** | 0.564 | | 0.366 | | 0.365 | **0.333** | 0.586 | | 0.390 | | 0.349 |
| ETTm2 | CRPS | **0.228** | 0.333 | | 0.247 | | 0.241 | **0.229** | 0.341 | | 0.248 | | 0.243 |
| Weather | CRPS | **0.226** | 0.329 | | 0.247 | | 0.246 | **0.233** | 0.336 | | 0.251 | | 0.250 |
| Exch. | CRPS | **0.340** | 0.374 | | 0.549 | | 0.457 | **0.341** | 0.385 | | 0.563 | | 0.462 |

Table 1: Four-horizon average results of uncertainty estimation in LTSF. The best results of IS90 and CRPS are highlighted in bold. For IS and CRPS, lower is better, while Cov90 should be interpreted relative to the nominal level of 0.90.

**Point Forecasting Results.** Table 2 reports MSE and MAE averaged uniformly across the four forecasting horizons. The full per-horizon results are reported in Appendix C3.We report both base retrieval embeddings and their late-fusion variant, Ours-Stack, which combines the complementary handcrafted and RFF forecasts under the same input-length protocol. We also tested the results of our method under multi-input-length fusion, but for the sake of fair comparison, we did not include it in the main table. Relevant results can be found in Appendix C4.

The proposed retrieval method outperforms all selected trained baselines on two datasets. On ETTh2, Ours-RFF achieves the best average result, with 0.299 MSE and 0.370 MAE, improving over TimeMixer++ (0.339/0.380). On ETTm2, Ours-Stack obtains 0.218 MSE and 0.311 MAE, compared with 0.269 for TimeMixer++ in MSE and 0.316 for CGTFra in MAE; the handcrafted component already reaches 0.218/0.312, while stacking slightly improves MAE. The fusion also improves or matches the stronger individual embedding on ETTh1, Weather, and Exchange, although trained models remain superior overall on ETTh1, ETTm1, Weather, and Exchange.

**Ablation Study.** At H=96 in uncertainty estimation, Lookback-PIT generally produces sharper candidate intervals, while validation-selected rate tempering determines whether this contraction should be retained or reversed. PIT has only a marginal effect on CRPS, confirming that full-distribution performance primarily originates from the retrieved empirical support. Freezing the retrieval archive produces little change in point-forecast accuracy on ETTh2 and ETTm2, indicating that their gains do not arise from access to newly observed test futures. Detailed ablation and archive-update results are reported in Appendix C5.

| Dataset | Ours-HC | Ours-RFF | Ours-Stack | CGTFra | TimeMixer++ | FilterTS | iTrans | PatchTST |
|---|---|---|---|---|---|---|---|---|
| | Mse/Mae | Mse/Mae | Mse/Mae | Mse/Mae | Mse/Mae | Mse/Mae | Mse/Mae | Mse/Mae |
| ETTh1 | 0.759/0.627 | 0.550/0.534 | 0.549/0.532 | 0.434/**0.427** | **0.418**/0.432 | 0.433/**0.430** | 0.454/0.467 | 0.507/0.472 |
| ETTh2 | 0.330/0.393 | **0.299/0.370** | **0.299**/0.371 | 0.369/0.394 | 0.339/0.380 | 0.372/0.396 | 0.383/0.407 | 0.391/0.411 |
| ETTm1 | 0.520/0.492 | 0.465/0.461 | 0.509/0.480 | 0.388/0.386 | **0.368/0.378** | 0.388/0.398 | 0.409/0.410 | 0.402/0.406 |
| ETTm2 | **0.218**/0.312 | 0.247/0.325 | **0.218/0.311** | 0.277/0.316 | 0.269/0.320 | 0.276/0.337 | 0.288/0.332 | 0.290/0.334 |
| Weather | 0.344/0.329 | 0.361/0.339 | 0.342/0.325 | 0.251/0.273 | **0.226/0.262** | 0.244/0.274 | 0.258/0.278 | 0.265/0.285 |
| Exch. | 0.493/0.459 | 0.472/0.455 | 0.427/0.440 | **0.318/0.384** | 0.357/0.409 | 0.352/0.397 | 0.360/0.403 | 0.366/0.404 |

Table 2: Four-horizon average point-forecasting results in LTSF. The best results of MSE and MAE are highlighted in bold.

**Retrieval Headroom Analysis.** To distinguish the limitations of KReF's whole-trajectory routing from the information available in the historical archive, we conduct an archive single-neighbor oracle analysis at H=96. At each forecast origin, the archive contains the training and validation samples together with only earlier test futures that have already been fully observed. Using the query future solely for ex-post selection, the oracle copies the MSE-nearest historical block, selecting one sample for the complete trajectory, independently per horizon, or independently per channel; no averaging or fitting is performed. Across six datasets, trajectory routing yields a mean MSE/MAE of 0.1954/0.2940. Horizon-wise routing reduces these errors to 0.0302/0.1096, corresponding to an 84.5% MSE reduction, while channel-wise routing obtains 0.0866/0.1786, a 55.7% MSE reduction. Horizon-wise MSE decreases by 93.2%–94.2% on the four ETT datasets, whereas Exchange favors channel-wise routing, which reduces MSE by 82.5%. Although non-deployable and not a strict upper bound, this analysis shows that finer routing can expose historical information hidden by whole-trajectory matching and motivates blockwise and learned retrieval. Full per-dataset results and implementation details are provided in Appendix E.

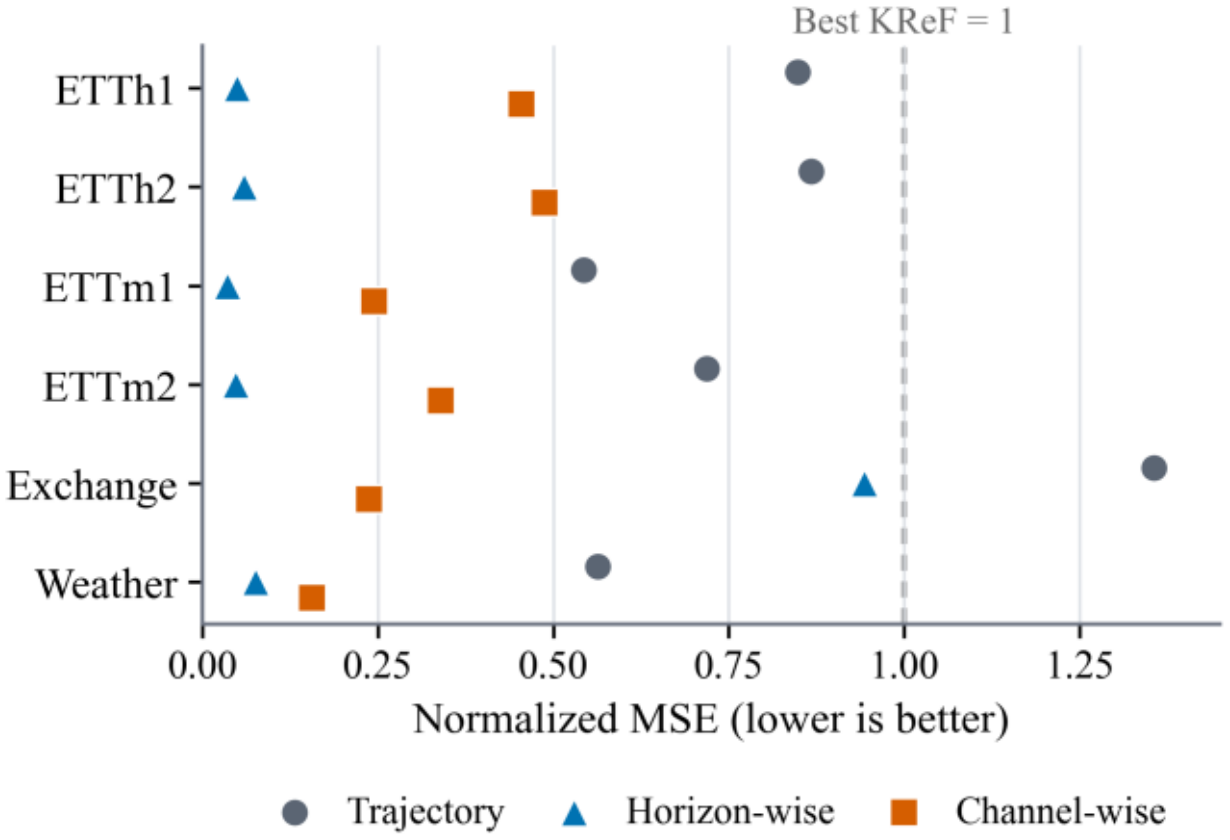

Figure 2: Retrieval headroom under archive single-neighbor oracle routing at H=96. MSE is normalized by the best reported KReF result on each dataset, shown by the dashed line at one; lower is better. Finer horizon- or channel-wise routing reveals substantial unused historical information.

## 5. Discussion

KReF is more broadly effective for uncertainty estimation than for point forecasting. Appendix D further analyzes the characteristics of the six datasets and demonstrates KReF's capability for uncertainty estimation under strong distributional shifts; retrieved futures can still characterize local dispersion when their mean is not the strongest point forecast. Lookback-PIT and expansion/shrinkage rates further adjust its distribution and interval, avoiding the reliance on coverage feedback required by online calibration controllers and the poor IS90 caused by delayed feedback. This helps explain its strong CRPS and IS90 results, particularly at long horizons, although lookback-to-future rank transfer may not provide a formal coverage guarantee.

The main limitations are computational scale and historical transferability. Storing lookback–future pairs and aggregating neighbor futures incurs memory costs that grow with dimensionality and prediction horizon, currently preventing long-horizon experiments on ECL and Traffic. Strong distribution shift can also reduce point accuracy and interval sharpness, while the absence of learned representations limits cross-dataset transfer.

These limitations suggest scalable and hybrid extensions, including clustering or approximate disk-backed indices, retrieval in decomposed or learned latent spaces, and blockwise, channel-adaptive, or autoregressive routing motivated by the oracle analysis. Combining retrieval with pretrained representations and developing diagnostics for when historical reuse remains reliable are promising directions.

## 6. Conclusion

We proposed a training-free retrieval framework for long-term time-series forecasting and uncertainty estimation. Retrieved historical futures support point forecasting and define a local empirical predictive distribution without a trained probabilistic head. Lookback-conditioned PIT reweighting adapts this distribution, while validation-selected rates adjust the reported prediction intervals. The archive-oracle analysis further reveals substantial headroom for finer-grained retrieval.

## Appendices

**Appendix Overview.** This supplement provides implementation details, complete horizon-wise results, applicability analyses, and additional theoretical interpretation. In particular, Appendix F establishes two properties of Lookback-PIT: its interval boundaries are equivalent to empirical probability-level selection under the standard left generalized inverse, and directional expansion and shrinkage can strictly reduce oracle boundary residual when inward and outward rank transfer exhibit different reliability.

**Clarifications and Typographical Corrections to the Main Text.** For precision, we clarify four points in the main text. These corrections concern notation and implementation details and do not alter any reported result or conclusion.

**1. Empirical CRPS.** In the empirical CRPS formula in the Method section, the pairwise-distance term contains a typographical indexing error. Its distance factor should be $\tilde{Y}_{\mathrm{j},h,c} - \tilde{Y}_{k,h,c}$, rather than $\tilde{Y}_{i,h,c} - \tilde{Y}_{k,h,c}$. The intended formula is: $CRPS\left(F^{\mathrm{LB}}_{i,h,c}, \tilde{Y}_{i,h,c}\right) = \sum_j \tilde{w}_{ij,h,c}\left|\tilde{Y}_{j,h,c} - \tilde{Y}_{i,h,c}\right| - \frac{1}{2}\sum_{j,k} \tilde{w}_{ij,h,c}\tilde{w}_{ik,h,c}\left|\tilde{Y}_{\mathrm{j},h,c} - \tilde{Y}_{k,h,c}\right|$. The implementation and all reported CRPS results use the correct pairwise term $\tilde{Y}_{\mathrm{j},h,c} - \tilde{Y}_{k,h,c}$.

**2. Shrinkage-rate notation.** In the Experimental Setup, the shrinkage rate is inadvertently denoted by $\eta_{\mathrm{e}}$. It should be denoted by $\eta_s$; $\eta_{\mathrm{e}}$ denotes the expansion rate. This is a notation-only correction, and the implementation uses separate expansion and shrinkage rates.

**3. Endpoint completion and quantile interpolation.** The main Method section omits one finite-sample implementation conventions used by Lookback-PIT. At the cumulative-mass endpoint, we set the transformed cumulative masses to $C_{i,K,h,c} = 1$, while the empirical mid-CDF transformation is applied at the interior cumulative-mass knots. Prediction boundaries are obtained by linearly interpolating between adjacent retrieved support values. These conventions are used in all experiments and are formally specified in Appendix F.1.

**4. Meaning of training-free uncertainty estimation.** The statement that the uncertainty baselines require neither a loss function nor iteration should be interpreted as stating that they do not require gradient-based training of a parametric predictive model. KReF and the uncertainty baselines are not hyperparameter-free: they use validation-based selection of their respective hyperparameters, as described in the Experimental Setup and Appendix A. Their online variants may also update calibration states sequentially, but none trains a parametric forecasting model through gradient backpropagation.

**Appendix-A: Interval baselines.** All interval methods use the same online-pessimistic KNN point forecasts. For multistep multivariate prediction, each method maintains a separate state for every horizon-channel coordinate. Feedback for a forecast issued at origin $i$ and lead $h$ is released only when its target becomes observable, preventing future leakage. Static Split Conformal estimates a finite-sample absolute-residual quantile from validation data. Online Split Conformal maintains a causally expanding residual pool and inserts each test residual only after the corresponding horizon-specific target has arrived.

MS-ACI applies the published multistep diagonal-feedback construction with the practical largest-observed-score clipping rule. We select $\eta$ from {0.1,0.05,0.01,0.005}. MS-C-PI uses the scorecaster-free P+I controller, searching $\eta \in \{1,0.5,0.1,0.05\}$, $(C_{sat}, K_I) \in \{(1,200),(1,100),(5,10)\}$, with a score-range window of 100. MS-COP uses its ECDF correction with window 100, $\lambda_t/\eta_t = 0.5$, and $\eta \in \{1,0.5,0.1,0.05\}$. For each method, one global configuration is selected using causal IS90 on a chronological validation suffix and is then frozen throughout testing; adaptive states remain horizon-channel specific.

**Distributional Baselines.** Gaussian-Res centers a Gaussian distribution at the shared KNN point forecast and estimates its scale separately for every horizon-channel coordinate. ER-Static treats signed validation residual trajectories as equally weighted empirical samples. ER-Expanding begins with causally observable validation residuals and inserts each test residual after its target becomes available, without discarding earlier residuals. Each residual trajectory remains an empirical sample rather than being averaged.

**Appendix-B: Uncertainty Metrics.** All metrics are averaged over valid (sample, horizon, channel) elements at nominal level $\alpha$ ($\alpha$=0.9 in the main text).

**Coverage** is the empirical fraction of targets inside the interval, $cov$ = mean $1[lo \le y \le hi]$, and should match $\alpha$.

**Interval score** (Gneiting and Raftery 2007) combines sharpness and calibration: $IS = (hi - lo) + (\frac{2}{1-\alpha} * [(lo - y)_+ + (y - hi)_+]$, rewarding narrow intervals while penalizing targets that fall outside; we report its mean.

**CRPS** scores full predictive distribution $F$: $CRPS(F, y) = \int (F(z) - 1[z \ge y])^2 dz$.

We therefore report CRPS only for KReF, Gaussian-Res, and the empirical-residual distributions. Split Conformal, Online Split Conformal, MS-ACI, MS-C-PI, and MS-COP specify interval boundaries but do not natively assign probability mass over possible outcomes. KReF computes CRPS from its PIT-reweighted retrieved futures; Gaussian-Res uses its fitted Gaussian distribution; ER-Static and ER-Expanding use their corresponding empirical residual distributions.

**Appendix-C1: Per-horizon interval quality.** Table 3 reports Cov90 and IS90 separately at every prediction horizon. KReF achieves the lowest CRPS in all 48 dataset-embedding-horizon settings, showing that its average CRPS advantage is not driven by a small number of favorable horizons. For IS90, KReF is strictly best in 34 settings and tied for best in one additional setting.

Interval coverage is less uniform, ranging from 0.783 (Exchange RFF on Horizon 720) to 0.917 (ETTh2 RFF on Horizon 720) across the 48 KReF configurations. This distinction is important: a lower CRPS indicates a better complete predictive distribution but does not imply exact calibration at every individual nominal level, while the expansion/shrinkage rate hyperparameters selected on the validation set to minimize the coverage loss rate may fail due to validation-test set drift, resulting in over-coverage/under-coverage. At H=720 on Exchange, all methods under-cover, yet KReF retains Cov90 of 0.845 and 0.783 under handcrafted and RFF embeddings, compared with 0.742 and 0.681 for MS-ACI. KReF also obtains the lowest IS90 in both settings. In contrast, the adaptive controllers often attain near-nominal average coverage elsewhere only through substantially expanded intervals, illustrating the difficulty of horizon-delayed feedback in LTSF.

| Emb | Dataset | H | Cov90 | | | | | | IS90 | | | | | |
|---|---|---|---|---|---|---|---|---|---|---|---|---|---|---|
| | | | Ours | SpC | SpC-OL | MS-ACI | MS-C-PI | MS-COP | Ours | SpC | SpC-OL | MS-ACI | MS-C-PI | MS-COP |
| Handcrafted | ETTh1 | 96 | 0.895 | 0.900 | 0.905 | 0.903 | 0.901 | 0.901 | **3.012** | 3.298 | 3.303 | 3.709 | 4.996 | 4.708 |
| | | 192 | 0.891 | 0.892 | 0.902 | 0.902 | 0.897 | 0.900 | **3.209** | 3.518 | 3.511 | 4.682 | 10.048 | 9.159 |
| | | 336 | 0.885 | 0.877 | 0.895 | 0.901 | 0.891 | 0.892 | **3.435** | 3.757 | 3.727 | 5.909 | 19.754 | 16.862 |
| | | 720 | 0.875 | 0.854 | 0.882 | 0.894 | 0.869 | 0.873 | **4.071** | 4.385 | 4.309 | 8.335 | 34.114 | 30.753 |
| | ETTh2 | 96 | 0.898 | 0.912 | 0.915 | 0.903 | 0.903 | 0.905 | **2.075** | 2.294 | 2.243 | 3.431 | 4.031 | 4.183 |
| | | 192 | 0.903 | 0.912 | 0.914 | 0.902 | 0.903 | 0.904 | **2.379** | 2.529 | 2.498 | 4.220 | 7.043 | 6.524 |
| | | 336 | 0.906 | 0.906 | 0.909 | 0.899 | 0.894 | 0.896 | 2.684 | 2.677 | **2.659** | 5.267 | 11.860 | 10.599 |
| | | 720 | 0.913 | 0.904 | 0.912 | 0.899 | 0.867 | 0.869 | 3.162 | 3.198 | **3.157** | 6.389 | 19.137 | 17.819 |
| | ETTm1 | 96 | 0.887 | 0.911 | 0.912 | 0.900 | 0.926 | 0.926 | **2.623** | 3.012 | 2.994 | 16.052 | 37.625 | 37.362 |
| | | 192 | 0.884 | 0.910 | 0.911 | 0.900 | 0.926 | 0.926 | **2.887** | 3.303 | 3.290 | 29.431 | 41.160 | 41.580 |
| | | 336 | 0.885 | 0.905 | 0.908 | 0.900 | 0.918 | 0.918 | **3.151** | 3.567 | 3.561 | 36.857 | 46.301 | 47.221 |
| | | 720 | 0.880 | 0.897 | 0.903 | 0.900 | 0.902 | 0.902 | **3.420** | 3.902 | 3.866 | 50.759 | 79.603 | 70.123 |
| | ETTm2 | 96 | 0.893 | 0.890 | 0.901 | 0.900 | 0.919 | 0.918 | **1.741** | 2.061 | 1.985 | 5.405 | 7.357 | 6.159 |
| | | 192 | 0.900 | 0.899 | 0.907 | 0.900 | 0.919 | 0.921 | **2.001** | 2.316 | 2.219 | 8.029 | 11.954 | 11.796 |
| | | 336 | 0.907 | 0.903 | 0.910 | 0.900 | 0.917 | 0.918 | **2.307** | 2.556 | 2.451 | 11.410 | 20.393 | 20.349 |
| | | 720 | 0.910 | 0.905 | 0.912 | 0.899 | 0.909 | 0.910 | **2.734** | 2.894 | 2.836 | 20.040 | 51.568 | 50.721 |
| | Weather | 96 | 0.888 | 0.852 | 0.875 | 0.900 | 0.916 | 0.916 | 2.177 | 1.957 | **1.936** | 6.290 | 6.911 | 6.739 |
| | | 192 | 0.879 | 0.848 | 0.872 | 0.899 | 0.915 | 0.915 | 2.529 | 2.367 | **2.325** | 8.922 | 12.788 | 12.941 |
| | | 336 | 0.885 | 0.845 | 0.870 | 0.898 | 0.908 | 0.909 | 2.977 | 2.727 | **2.667** | 12.445 | 22.178 | 21.194 |
| | | 720 | 0.890 | 0.838 | 0.868 | 0.894 | 0.891 | 0.893 | 3.268 | 3.175 | **3.143** | 16.972 | 34.542 | 33.318 |
| | Exchan | 96 | 0.878 | 0.835 | 0.871 | 0.894 | 0.906 | 0.907 | **1.597** | 1.597 | 1.663 | 2.761 | 4.580 | 4.423 |
| | | 192 | 0.851 | 0.797 | 0.849 | 0.885 | 0.895 | 0.898 | **2.363** | 2.493 | 2.590 | 4.114 | 7.385 | 7.293 |
| | | 336 | 0.832 | 0.741 | 0.812 | 0.852 | 0.866 | 0.869 | **3.258** | 3.801 | 3.791 | 5.113 | 13.110 | 12.466 |
| | | 720 | 0.845 | 0.673 | 0.711 | 0.742 | 0.705 | 0.706 | **6.765** | 7.088 | 6.997 | 7.162 | 11.713 | 11.322 |
| Random Fourier Features | ETTh1 | 96 | 0.891 | 0.907 | 0.909 | 0.903 | 0.901 | 0.901 | **2.506** | 2.909 | 2.907 | 3.335 | 4.312 | 4.070 |
| | | 192 | 0.893 | 0.897 | 0.907 | 0.903 | 0.893 | 0.898 | **2.745** | 3.139 | 3.119 | 4.200 | 8.471 | 7.765 |
| | | 336 | 0.890 | 0.881 | 0.900 | 0.903 | 0.886 | 0.890 | **3.016** | 3.400 | 3.353 | 5.212 | 16.297 | 14.621 |
| | | 720 | 0.873 | 0.860 | 0.890 | 0.897 | 0.865 | 0.868 | **3.645** | 3.991 | 3.895 | 7.154 | 28.524 | 25.335 |
| | ETTh2 | 96 | 0.903 | 0.914 | 0.917 | 0.903 | 0.903 | 0.905 | **2.028** | 2.274 | 2.218 | 3.460 | 4.055 | 4.244 |
| | | 192 | 0.903 | 0.912 | 0.914 | 0.903 | 0.901 | 0.903 | **2.288** | 2.511 | 2.477 | 4.082 | 6.771 | 6.351 |
| | | 336 | 0.904 | 0.908 | 0.910 | 0.899 | 0.893 | 0.895 | **2.524** | 2.646 | 2.625 | 4.911 | 10.778 | 9.617 |
| | | 720 | 0.917 | 0.911 | 0.916 | 0.900 | 0.865 | 0.867 | **2.982** | 3.125 | 3.084 | 6.061 | 17.570 | 16.296 |
| | ETTm1 | 96 | 0.871 | 0.914 | 0.913 | 0.901 | 0.927 | 0.926 | **2.493** | 2.967 | 2.943 | 15.077 | 37.943 | 37.650 |
| | | 192 | 0.870 | 0.910 | 0.911 | 0.900 | 0.928 | 0.928 | **2.637** | 3.237 | 3.220 | 28.349 | 44.021 | 40.935 |
| | | 336 | 0.870 | 0.905 | 0.907 | 0.900 | 0.918 | 0.919 | **2.884** | 3.487 | 3.477 | 37.172 | 48.599 | 54.028 |
| | | 720 | 0.862 | 0.896 | 0.902 | 0.900 | 0.903 | 0.904 | **3.151** | 3.806 | 3.766 | 50.154 | 79.769 | 64.801 |
| | ETTm2 | 96 | 0.887 | 0.894 | 0.902 | 0.900 | 0.919 | 0.919 | **1.815** | 2.086 | 2.017 | 5.297 | 6.948 | 5.801 |
| | | 192 | 0.900 | 0.912 | 0.914 | 0.900 | 0.918 | 0.921 | 2.287 | 2.360 | **2.279** | 7.596 | 11.480 | 11.191 |
| | | 336 | 0.906 | 0.921 | 0.920 | 0.900 | 0.918 | 0.918 | 2.637 | 2.603 | **2.520** | 10.648 | 20.848 | 20.980 |
| | | 720 | 0.914 | 0.920 | 0.919 | 0.899 | 0.909 | 0.911 | 3.147 | 2.935 | **2.888** | 19.005 | 48.878 | 48.708 |
| | Weather | 96 | 0.871 | 0.849 | 0.874 | 0.900 | 0.919 | 0.919 | 2.100 | 2.031 | **1.992** | 6.366 | 7.874 | 7.483 |
| | | 192 | 0.864 | 0.847 | 0.873 | 0.899 | 0.916 | 0.916 | 2.527 | 2.389 | **2.338** | 8.598 | 14.285 | 13.471 |
| | | 336 | 0.871 | 0.844 | 0.871 | 0.898 | 0.909 | 0.910 | **2.634** | 2.764 | 2.679 | 13.020 | 25.014 | 23.151 |
| | | 720 | 0.885 | 0.837 | 0.868 | 0.894 | 0.893 | 0.894 | 3.688 | 3.227 | **3.167** | 17.613 | 36.635 | 35.269 |
| | Exchan | 96 | 0.865 | 0.833 | 0.867 | 0.894 | 0.906 | 0.907 | 1.711 | **1.681** | 1.738 | 2.900 | 4.587 | 4.474 |
| | | 192 | 0.831 | 0.796 | 0.849 | 0.886 | 0.889 | 0.892 | 2.337 | **2.600** | 2.647 | 4.195 | 7.832 | 7.869 |
| | | 336 | 0.816 | 0.741 | 0.817 | 0.854 | 0.856 | 0.859 | 3.592 | 3.916 | **3.792** | 5.041 | 12.629 | 11.948 |
| | | 720 | 0.783 | 0.587 | 0.654 | 0.681 | 0.677 | 0.676 | 6.240 | 8.313 | **7.750** | 7.931 | 11.701 | 11.303 |

Table 3: Per-horizon uncertainty-estimation results in LTSF. The best results of IS90 are highlighted in bold.

**Appendix-C2:** Table 4 reports four-horizon-average coverage at nominal levels of 50% and 80%. Cov50 ranges from 0.421 to 0.505, while Cov80 ranges from 0.730 to 0.806. Thus, several central intervals are systematically under-covered even though the 90% intervals are generally better calibrated. Validation selection frequently reaches the maximum expansion candidate of 3.0 at the 50% level, indicating that the compact search grid constrains some interval adjustments.

| Dataset | HC Cov50 | HC Cov80 | RFF Cov50 | RFF Cov80 |
|---|---|---|---|---|
| ETTh1 | 0.4748 | 0.7741 | 0.4210 | 0.7688 |
| ETTh2 | 0.5049 | 0.7993 | 0.4749 | 0.8058 |
| ETTm1 | 0.4764 | 0.7791 | 0.4598 | 0.7533 |
| ETTm2 | 0.4623 | 0.7903 | 0.4607 | 0.7956 |
| Exchange | 0.4740 | 0.7362 | 0.4609 | 0.7297 |
| Weather | 0.4982 | 0.7882 | 0.4887 | 0.7807 |

Table 4: Four-horizon average coverage results at the 50% and 80% nominal levels.

We therefore conduct a uniform post-hoc sensitivity experiment over all six datasets and both embeddings. The expansion candidate grid is extended to $\{1, 2, 3, 5, 8, \ldots, 2584\}$, while the shrinkage grid remains unchanged. Because an unrestricted expansion grid can produce extreme sample-level intervals, we apply the same trust-region factor $\kappa$ =8 to every dataset, embedding, horizon, and nominal level. The expanded grid and clipping factor participate in validation selection and are then frozen during testing. These settings are used only for sensitivity analysis and do not replace the compact-grid main results.

As shown in Table 5, the expanded configuration changes the central-interval calibration pattern substantially. Average Cov50 increases from 0.471 to 0.508, eliminating several severe under-coverage cases, although Exchange-RFF becomes over-covered at 0.610. Consequently, the mean absolute Cov50 deviation changes only slightly, from 0.029 to 0.032. At the 80% level, average coverage improves from 0.775 to 0.795 and the mean absolute deviation decreases from 0.026 to 0.024. The improvement is especially clear on Exchange, where Cov80 increases from 0.736/0.730 to 0.789/0.787 under handcrafted/RFF embeddings. In contrast, ETTh2 becomes over-covered at 0.848/0.842, illustrating that greater expansion capacity does not transfer uniformly. The fixed trust region keeps the resulting widths finite: Width50 ranges from 0.505 to 1.955 and Width80 from 1.259 to 4.226. This avoids the pathological widths observed under unrestricted expansion, but the comparatively wide ETTh2 and Exchange intervals show that clipping cannot by itself resolve validation-to-test mismatch.

| Datasets | Emb | Cov50 | Wid50 | Cov80 | Wid80 |
|---|---|---|---|---|---|
| ETTh1 | HC | 0.4748 | 1.0152 | 0.7741 | 1.9562 |
| | RFF | 0.4871 | 1.4621 | 0.7759 | 2.4339 |
| ETTh2 | HC | 0.5268 | 0.6899 | 0.8480 | 3.0826 |
| | RFF | 0.5184 | 0.7141 | 0.8417 | 2.7728 |
| ETTm1 | HC | 0.4764 | 0.7683 | 0.7774 | 2.1054 |
| | RFF | 0.4652 | 0.6964 | 0.7434 | 1.6440 |
| ETTm2 | HC | 0.5372 | 0.7578 | 0.8193 | 2.4481 |
| | RFF | 0.4665 | 0.5051 | 0.8081 | 1.9612 |
| Exch. | HC | 0.5469 | 1.6147 | 0.7888 | 4.2264 |
| | RFF | 0.6098 | 1.9551 | 0.7872 | 4.1625 |
| Weather | HC | 0.5030 | 0.5140 | 0.7978 | 1.9220 |
| | RFF | 0.4875 | 0.5148 | 0.7798 | 1.2594 |

Table 5: Four-horizon-average sensitivity results using an expansion-rate candidate grid extending to 2584 and a fixed trust-region factor $\kappa$ =8.

It should be noted that although the experiments in Appendix C2 include 85 combinations of expansion and contraction rates, the overall hyperparameter grid search is actually very fast because the intermediate results of the computation on the validation set can be saved. Besides, the finite rate grid is an implementation choice rather than an intrinsic requirement of KReF. An exact breakpoint- or order-statistic-based selector could therefore replace the manually specified grid without repeating retrieval. Exact validation optimization, however, is not guaranteed to transfer better under validation–test shift. In the expanded-grid sensitivity experiment, the additional expansion capacity changes Exchange-RFF Cov50 from 0.461 to 0.610. The result does not establish that an exact selector necessarily overfits, but they demonstrate that greater validation-time flexibility can amplify calibration mismatch on the test distribution. We therefore retain a coarse, reliability-constrained rate grid as a form of regularization rather than pursuing the exact empirical validation optimum.

A further extension could allow signed rates, which reflect an inward PIT proposal into an outward correction, or vice versa. Such a generalization would require explicit non-crossing constraints and is left for future work.

**Appendix-C3: Per-Horizon Point-Forecasting Results.** Table 6 reports MSE and MAE separately at each prediction horizon. The four-horizon averages in the main text do not arise from an isolated short-horizon gain. On ETTh2, the KReF variants achieve the lowest MSE at all four horizons: Ours-Stack obtains 0.232 at H=96, Ours-RFF/Stack obtain 0.278 at H=192, Ours-RFF obtains 0.307 at H=336, and Ours-RFF/Stack obtain 0.378 at H=720. They also achieve the lowest MAE through H=336, while TimeMixer++ is slightly better in MAE at H=720.

On ETTm2, Ours-Stack or Ours-HC attains the lowest MSE at every horizon, from 0.158 at H=96 to 0.294 at H=720; Ours-Stack further achieves the lowest MAE at the longer horizons H=336 and H=720. Thus, the main-table gains on ETTh2 and ETTm2 persist across forecasting lengths, especially in MSE. On the remaining datasets, the per-horizon results show that trained baselines outperform KReF at most horizons, confirming that KReF gains are dataset-dependent rather than being obscured by horizon averaging.

| Dataset | H | Ours-HC | Ours-RFF | Ours-Stack | CGTFra | TimeMixer++ | FilterTS | iTrans | PatchTST |
|---|---|---|---|---|---|---|---|---|---|
| Metrics | | Mse/Mae | Mse/Mae | Mse/Mae | Mse/Mae | Mse/Mae | Mse/Mae | Mse/Mae | Mse/Mae |
| ETTh1 | 96 | 0.648/0.561 | 0.443/0.465 | 0.439/0.461 | 0.372/**0.387** | **0.361**/0.403 | 0.374/0.391 | 0.386/0.405 | 0.460/0.447 |
| | 192 | 0.708/0.599 | 0.497/0.503 | 0.493/0.498 | 0.424/**0.418** | **0.416**/0.441 | 0.424/0.421 | 0.441/0.512 | 0.477/0.429 |
| | 336 | 0.764/0.632 | 0.554/0.538 | 0.551/0.534 | 0.473/0.443 | **0.430/0.434** | 0.464/0.441 | 0.487/0.458 | 0.546/0.496 |
| | 720 | 0.917/0.716 | 0.705/0.632 | 0.713/0.637 | 0.473/0.464 | **0.467/0.451** | 0.470/0.466 | 0.503/0.491 | 0.544/0.517 |
| ETTh2 | 96 | 0.246/0.339 | 0.235/0.326 | **0.232/0.324** | 0.288/0.336 | 0.276/0.328 | 0.290/0.338 | 0.297/0.349 | 0.308/0.355 |
| | 192 | 0.296/0.373 | **0.278/0.356** | **0.278/0.356** | 0.364/0.384 | 0.342/0.379 | 0.374/0.390 | 0.380/0.400 | 0.393/0.405 |
| | 336 | 0.333/0.398 | **0.307/0.376** | 0.310/0.379 | 0.410/0.422 | 0.346/0.398 | 0.406/0.420 | 0.428/0.432 | 0.427/0.436 |
| | 720 | 0.446/0.461 | **0.378**/0.422 | **0.378**/0.424 | 0.414/0.433 | 0.392/**0.415** | 0.418/0.437 | 0.427/0.445 | 0.436/0.450 |
| ETTm1 | 96 | 0.421/0.437 | 0.361/0.396 | 0.406/0.424 | 0.315/0.344 | **0.310/0.334** | 0.321/0.360 | 0.334/0.368 | 0.352/0.374 |
| | 192 | 0.484/0.470 | 0.437/0.446 | 0.492/0.469 | 0.366/0.372 | **0.348/0.362** | 0.363/0.382 | 0.390/0.393 | 0.374/0.387 |
| | 336 | 0.553/0.508 | 0.501/0.480 | 0.539/0.493 | 0.398/0.395 | **0.376/0.391** | 0.395/0.403 | 0.420/0.420 | 0.421/0.414 |
| | 720 | 0.621/0.552 | 0.563/0.522 | 0.600/0.532 | 0.472/0.435 | **0.440/0.423** | 0.474/0.446 | 0.491/0.459 | 0.462/0.449 |
| ETTm2 | 96 | **0.159**/0.266 | 0.165/0.268 | **0.158**/0.264 | 0.171/0.249 | 0.170/**0.245** | 0.172/0.255 | 0.180/0.264 | 0.183/0.270 |
| | 192 | **0.190**/0.293 | 0.218/0.305 | 0.190/0.293 | 0.238/0.293 | 0.229/**0.291** | 0.237/0.299 | 0.250/0.309 | 0.255/0.314 |
| | 336 | **0.229/0.322** | 0.272/0.341 | 0.229/**0.322** | 0.300/0.333 | 0.303/0.343 | 0.299/0.398 | 0.311/0.348 | 0.309/0.347 |
| | 720 | 0.295/0.368 | 0.333/0.384 | **0.294/0.367** | 0.397/0.391 | 0.373/0.399 | 0.397/0.394 | 0.412/0.407 | 0.412/0.404 |
| Weather | 96 | 0.242/0.258 | 0.263/0.274 | 0.239/0.257 | **0.152/0.190** | 0.155/0.205 | 0.162/0.207 | 0.174/0.214 | 0.186/0.227 |
| | 192 | 0.317/0.305 | 0.317/0.317 | 0.309/0.302 | 0.203/**0.239** | **0.201**/0.245 | 0.209/0.252 | 0.221/0.254 | 0.234/0.265 |
| | 336 | 0.366/0.347 | 0.384/0.353 | 0.359/0.344 | 0.257/0.279 | **0.237/0.265** | 0.263/0.294 | 0.278/0.296 | 0.284/0.301 |
| | 720 | 0.451/0.405 | 0.478/0.411 | 0.459/0.398 | 0.338/**0.334** | **0.312/0.334** | 0.344/0.344 | 0.358/0.347 | 0.356/0.349 |
| Exchange | 96 | 0.117/0.231 | 0.119/0.241 | 0.114/0.231 | **0.083/0.202** | 0.085/0.214 | **0.081/0.199** | 0.086/0.206 | 0.088/0.205 |
| | 192 | 0.208/0.322 | 0.215/0.327 | 0.204/0.318 | **0.173/0.296** | 0.175/0.313 | **0.171/0.294** | 0.177/0.299 | 0.176/0.299 |
| | 336 | 0.393/0.452 | 0.401/0.457 | 0.388/0.450 | 0.324/0.412 | 0.316/0.420 | 0.321/0.409 | 0.331/0.417 | **0.301/0.397** |
| | 720 | 1.255/0.832 | 1.154/0.795 | 1.001/0.759 | **0.668/0.619** | 0.851/0.689 | 0.837/0.688 | 0.847/0.691 | 0.901/0.714 |

Table 6: Per-horizon point estimates in LTSF results. The best results are highlighted in bold.

**Appendix-C4: Multi-Sequence-Length Softmax.** We further evaluate a multi-sequence-length variant that aggregates predictions from lookback lengths {48, 96, 192, 336, 720}. As shown in Table 7, this variant improves 14/18 results at prediction length 96 and 14/18 results at prediction length 720. The gains are especially clear on Weather, where the stacked result improves from 0.239 to 0.185 at horizon 96. These results suggest that different lookback lengths capture complementary recurrence patterns, which can be exploited without training additional models. Exchange under long-horizon forecasting is the main exception, where multi-sequence aggregation degrades performance, especially after validation-fitted stacking. This indicates that adaptive lookback aggregation can be sensitive under distribution shift.

| Dataset | Embedding/model | Multi input 96 | Original 96 | Multi input 720 | Original 720 |
|---|---|---|---|---|---|
| ETT h1 | Handcrafted | **0.627** | 0.648 | **0.889** | 0.917 |
| | RFF | **0.437** | 0.443 | **0.677** | 0.706 |
| | Stacking | **0.439** | **0.439** | **0.661** | 0.713 |
| ETT h2 | Handcrafted | **0.239** | 0.246 | **0.420** | 0.447 |
| | RFF | **0.228** | 0.235 | **0.374** | 0.378 |
| | Stacking | 0.233 | **0.232** | 0.393 | **0.378** |
| ETT m1 | Handcrafted | **0.389** | 0.421 | **0.568** | 0.621 |
| | RFF | **0.338** | 0.362 | **0.495** | 0.563 |
| | Stacking | **0.335** | 0.406 | **0.497** | 0.600 |
| ETT m2 | Handcrafted | **0.153** | 0.159 | **0.274** | 0.295 |
| | RFF | **0.152** | 0.165 | **0.265** | 0.333 |
| | Stacking | **0.151** | 0.158 | **0.269** | 0.294 |
| Weather | Handcrafted | **0.186** | 0.242 | **0.353** | 0.462 |
| | RFF | **0.198** | 0.269 | **0.363** | 0.478 |
| | Stacking | **0.185** | 0.239 | **0.345** | 0.459 |
| Ex-chang | Handcrafted | 0.114 | **0.113** | 1.699 | **1.255** |
| | RFF | 0.130 | **0.119** | 1.859 | **1.154** |
| | Stacking | **0.110** | 0.114 | 4.479 | **1.005** |

Table 7: Point-forecasting results of multi-input-length aggregation at prediction horizons 96 and 720.

**Appendix-C5: Uncertainty Component Ablation.** We evaluate three nested variants at prediction length 96. Base-WQ directly forms intervals and predictive masses from the retrieval weights. Lookback-PIT uses the observed query lookback to produce query-specific probability reweighting and candidate bounds. Full additionally applies the validation-selected expansion and shrinkage rates to temper the PIT-induced bound displacement.

The two calibration components play complementary roles. Lookback-PIT generally produces sharper candidate intervals, reducing the average 90% width from 1.485 to 1.394, although this sharpening can also reduce coverage. Rate tempering determines how much of this contraction should be retained and how strongly outward adjustments should be amplified. Among the three configurations for which Base-WQ already attains Cov90 above 0.89, Full maintains approximately nominal coverage between 0.887 and 0.896 while reducing IS90 in all three cases, by 0.033 on average. Among the remaining nine initially under-covered configurations, Full increases Cov90 in every case, raising their average coverage from 0.866 to 0.884. Thus, the combined method preserves useful sharpening when the base distribution is already sufficiently calibrated, while preferentially widening intervals when the retrieved distribution is under-dispersed.

PIT reweighting has only a marginal effect on CRPS: its average value changes from 0.2406 for Base-WQ to 0.2410, with individual differences no larger than 0.0019. This indicates that KReF's CRPS advantage primarily originates from the locally retrieved empirical distribution. Lookback-PIT and rate tempering instead act mainly as query-specific interval adaptation mechanisms. Their heterogeneous effects are consistent with lookback-to-future rank transfer being an approximate, dataset-dependent relationship rather than a universal assumption.

| Dataset | Emb | Metric | Base | +PIT | Full |
|---|---|---|---|---|---|
| ETTh1 | HC | Cov90 | 0.894 | 0.876 | 0.896 |
| | | IS90 | 3.047 | 2.989 | 3.012 |
| | RFF | Cov90 | 0.899 | 0.839 | 0.891 |
| | | IS90 | 2.529 | 2.546 | 2.507 |
| ETTh2 | HC | Cov90 | 0.885 | 0.874 | 0.898 |
| | | IS90 | 2.075 | 2.078 | 2.075 |
| | RFF | Cov90 | 0.888 | 0.861 | 0.904 |
| | | IS90 | 2.021 | 1.996 | 2.028 |
| ETTm1 | HC | Cov90 | 0.905 | 0.863 | 0.887 |
| | | IS90 | 2.665 | 2.561 | 2.623 |
| | RFF | Cov90 | 0.838 | 0.817 | 0.871 |
| | | IS90 | 2.461 | 2.437 | 2.493 |
| ETTm2 | HC | Cov90 | 0.878 | 0.856 | 0.893 |
| | | IS90 | 1.723 | 1.742 | 1.741 |
| | RFF | Cov90 | 0.864 | 0.848 | 0.887 |
| | | IS90 | 1.754 | 1.779 | 1.815 |
| Exch. | HC | Cov90 | 0.876 | 0.878 | 0.878 |
| | | IS90 | 1.584 | 1.597 | 1.597 |
| | RFF | Cov90 | 0.856 | 0.851 | 0.865 |
| | | IS90 | 1.644 | 1.635 | 1.711 |
| Weather | HC | Cov90 | 0.872 | 0.871 | 0.889 |
| | | IS90 | 1.708 | 1.769 | 2.177 |
| | RFF | Cov90 | 0.837 | 0.845 | 0.871 |
| | | IS90 | 1.777 | 1.772 | 2.100 |

Table 8: Uncertainty-component ablation at H=96. Base denotes Base-WQ, +PIT adds Lookback-PIT, and Full additionally applies the interval adjustment.

**Effect of Test-Time Archive Updates.** The main experiments use the online-pessimistic protocol to align the point and uncertainty pipelines. To quantify the contribution of archive adaptation, we additionally evaluate a static offline variant whose retrieval database contains only training and validation instances and remains fixed throughout testing. As shown in Table 9, online updates have little effect on ETTh2 and ETTm2, where KReF obtains its strongest point-forecasting results. The larger improvement on ETTm1 does not change the method ranking. These results indicate that KReF's two favorable comparisons are primarily attributable to retrieval from pre-test history rather than access to newly observed test futures.

| Dataset | HC Online | HC Offline | ΔMSE | RFF Online | RFF Offline | ΔMSE |
|---|---|---|---|---|---|---|
| ETTh1 | 0.759/ 0.627 | 0.778/ 0.637 | +2.5% | 0.550/ 0.534 | 0.570/ 0.551 | +3.63% |
| ETTh2 | 0.330/ 0.393 | 0.334/ 0.395 | +1.21% | 0.299/ 0.370 | 0.302/ 0.373 | +1.0% |
| ETTm1 | 0.520/ 0.492 | 0.549/ 0.510 | +5.57% | 0.465/ 0.461 | 0.514/ 0.491 | +10.54% |
| ETTm2 | 0.218/ 0.312 | 0.219/ 0.313 | +0.46% | 0.247/ 0.325 | 0.249/ 0.326 | +0.81% |
| Weather | 0.344/ 0.329 | 0.348/ 0.333 | +1.16% | 0.361/ 0.339 | 0.361/ 0.341 | −0.0% |
| Exch | 0.493/ 0.459 | 0.494/ 0.461 | +0.02% | 0.472/ 0.455 | 0.472/ 0.455 | −0.0% |

Table 9: Online uses the pessimistic update protocol; Offline freezes the retrieval archive after the validation split. Each cell reports MSE/MAE averaged over four horizons.

**Appendix-D: What Makes KReF Work?** For each benchmark we measure STL seasonal strength (Cleveland et al. 1990), noise-to-signal ratio (Hyndman and Athanasopoulos 2018), the train–test Wasserstein-1 shift (Peyré and Cuturi 2019) and a similarity-rank ratio that splits each query's K neighbors into similarity quintiles and reports MSE(least-similar)/MSE(most-similar); a value below 1 means the most-similar neighbors are no better—slightly worse—than distant ones. (Exact definitions are provided in the code supplement.)

The point-forecasting results indicate that the applicability of direct retrieval depends jointly on predictability and temporal stability rather than on any single diagnostic. KReF performs particularly well on ETTh2 and ETTm2, which exhibit low residual noise-to-signal ratios and moderate distribution shift. Exchange is even cleaner, but combines the weakest seasonal structure with the largest shift, making historical futures less transferable despite favorable local similarity. These observations help explain why retrieval can be highly effective on selected datasets without yielding uniform improvements over trained forecasters.

The distributional results are substantially broader than the point-forecasting gains. KReF achieves the lowest CRPS across all evaluated datasets, embeddings, and horizons, including settings in which its retrieved mean is not the strongest point forecast. The component ablation shows that this advantage is already present in the base locally retrieved empirical distribution, while Lookback-PIT changes CRPS only marginally. Interval estimation exhibits a different pattern.

Across the six benchmarks, KReF's IS90 advantage does not vary monotonically with seasonality, noise-to-signal ratio, or similarity rank. It obtains the lowest average IS90 for both embeddings on ETTh1, ETTh2, ETTm1, and Exchange, and for the handcrafted embedding on ETTm2. These datasets span seasonal-strength values from 0.118 to 0.519, noise-to-signal ratios from 0.045 to 0.378, and similarity ranks from 0.814 to 1.058. This diversity suggests that interval quality is not determined solely by how accurately retrieved futures estimate the conditional mean.

Weather is the clearest exception. It has the smallest measured shift, 0.060, and Online Split Conformal achieves the best IS90 for both embeddings. Under such stable conditions, an expanding residual calibration pool transfers effectively to the test period and can produce particularly sharp intervals. Conversely, Exchange has the largest shift, 0.288. Residual-based and delayed adaptive calibration methods either under-cover at long horizons or substantially widen their intervals, whereas KReF retains comparatively strong coverage with sharper intervals. This contrast suggests that KReF is especially useful when globally accumulated residuals transfer poorly across time. Given the limited number of datasets, however, this should be interpreted as an empirical pattern rather than a causal conclusion.

| Dataset | Seasonal | Noise/signal | Shift ($W_1$) | Sim-rank HC | Sim-rank RFF |
|---|---|---|---|---|---|
| ETTh2 | 0.362 | 0.247 | 0.114 | 0.965 | 0.945 |
| ETTm2 | 0.376 | 0.243 | 0.111 | 0.916 | 0.843 |
| ETTh1 | 0.518 | 0.378 | 0.074 | 1.011 | 1.058 |
| ETTm1 | 0.519 | 0.377 | 0.072 | 0.963 | 1.005 |
| Weather | 0.402 | 0.475 | 0.060 | 0.901 | 0.939 |
| Exchange | 0.118 | 0.045 | 0.288 | 0.822 | 0.814 |

Table 10: Characteristics of the six benchmark datasets.

**Appendix-E: Oracle Analysis of Retrieval Headroom.** To distinguish the limitations of whole-future matching from the information available in the historical archive, we conduct a single-neighbor oracle analysis at $H$=96. At forecast origin i, the admissible archive $A_i$ contains the training and validation samples together with earlier test futures that have already been fully observed. Let $Y_i \in R^{H\times C}$ denote the benchmark-standardized query future, and let $B$ be a block of horizon-channel cells. The oracle selects

$$j_{i,B}^{*} = arg \min_{j\in A_i} \frac{1}{|B|} \sum_{(h,c)\in B} (Y_{i,h,c} - Y_{j,h,c})^2$$

and directly copies the corresponding block from that single historical sample:

$$\hat{Y}_{i,h,c} = Y_{j_{i,B}^{*},h,c}, \qquad (h,c) \in B$$

Trajectory routing uses one block containing all $H \times C$ cells and therefore selects one historical future for the complete prediction. Horizon-wise routing uses H blocks, selecting one sample independently for each horizon while jointly matching all channels. Channel-wise routing uses C blocks, selecting one complete H-step trajectory independently for each channel. No averaging, weighting, or fitting is performed.

The assembled predictions are then evaluated using exactly the same MSE and MAE as in the main experiments, averaged over all test samples, horizons, and channels. Both matching and evaluation are performed in the globally standardized benchmark space. The reported errors are therefore directly comparable to the point-forecasting results in the main experiments, although the oracle selection itself uses target-future information.

Table 11 reveals substantial information beyond whole-trajectory matching. Across the six datasets, horizon-wise routing reduces mean MSE from 0.1954 to 0.0302 and mean MAE from 0.2940 to 0.1096, corresponding to reductions of 84.5% and 62.7%. Channel-wise routing reduces them to 0.0866 and 0.1786. The four ETT datasets exhibit particularly consistent horizon-wise gains, with MSE reductions between 93.2% and 94.2%, indicating that their archives contain close analogues for individual future states even when no single historical continuation matches the complete horizon.

The preferred routing granularity is dataset dependent. Exchange benefits considerably more from channel-wise routing, which reduces MSE by 82.5%, whereas horizon-wise routing reduces it by only 30.4%. Weather benefits from both decompositions, attaining its lowest MSE under horizon-wise routing and its lowest MAE under channel-wise routing. These contrasting patterns suggest that reusable historical information may be organized along different temporal and variable-specific structures across datasets.

Because the true query future determines the selected samples, these oracle results are difficult to realize directly without an effective routing mechanism. However, they are not strict upper bounds on retrieval-based forecasting: the analysis is restricted to one historical sample per block and does not consider multi-neighbor aggregation, adaptive weighting, interpolation, local fitting, or learned retrieval. The results instead provide an optimistic reference for the information exposed by finer-grained routing and motivate blockwise, horizon-conditioned, channel-adaptive, and autoregressive retrieval beyond the current KReF design.

| Dataset | Metric | Trajectory | Horizon-wise | Channel-wise |
|---|---|---|---|---|
| ETTh1 | MSE | 0.3722 | 0.0217 | 0.1998 |
| | MAE | 0.4268 | 0.1095 | 0.3119 |
| ETTh2 | MSE | 0.2013 | 0.0137 | 0.1131 |
| | MAE | 0.3089 | 0.0816 | 0.2224 |
| ETTm1 | MSE | 0.1962 | 0.0128 | 0.0883 |
| | MAE | 0.3108 | 0.0840 | 0.2046 |
| ETTm2 | MSE | 0.1135 | 0.0075 | 0.0537 |
| | MAE | 0.2316 | 0.0600 | 0.1480 |
| Exchange | MSE | 0.1546 | 0.1076 | 0.0271 |
| | MAE | 0.2910 | 0.2427 | 0.1137 |
| Weather | MSE | 0.1347 | 0.0180 | 0.0374 |
| | MAE | 0.1946 | 0.0800 | 0.0708 |

Table 11: Archive-oracle errors under different routing granularities at H=96.

**Appendix-F: Boundary Analysis of Lookback-PIT and Directional Tempering.** This section gives a boundary-space interpretation of Lookback-PIT and explains why separate expansion and shrinkage rates can improve empirical tail calibration. We use one-dimensional notation for clarity; in the full multivariate setting, a single query-specific mid-CDF is constructed by pooling all valid lookback–channel coordinates and is then applied coordinatewise to every future cell.

**F.1. Equivalence between Lookback-PIT and empirical quantile selection.** For query *i*, let the retrieved future values at one coordinate be sorted as $Z_{i,(1)} \leq \ldots \leq Z_{i,(K)}$, with weights $W_{i,(k)}$. Their cumulative base masses are

$$C_{i,k} = \sum_{r=1}^{k} w_{i,(r)}\,, C_{i,0} = 0.$$

To match the finite-sample implementation at the endpoints, we use the endpoint-completed empirical mid-CDF. Applying the midrank definition for $0 < u < 1$ and setting $G_i(0) = 0$ and $G_i(1) = 1$. The endpoint completion ensures that the transformed masses form a proper probability distribution. Lookback-PIT transforms the cumulative future masses into $\tilde{C}_{i,k} = G_i(C_{i,k})$; consequently, the adjusted mass assigned to the *k-th* sorted future equals the increment between two adjacent transformed cumulative masses. For target probability *p*, the selected support index is

$$k_i^*(p) = min\{k: G_i(C_{i,k}) \geq p\}.$$

Equivalently, define the smallest admissible base probability level as

$$u_i^*(p) = min\{C_{i,k}: G_i(C_{i,k}) \geq p\}.$$

The resulting Lookback-PIT boundary therefore satisfies

$$Q_i^{LB}(p) = Q_i^B(u_i^*(p)).$$

In the continuous or interpolated representation, the same identity can be written using the left generalized inverse,

$$Q_i^{LB}(p) = Q_i^{\mathrm{B}}(G_i^{\leftarrow}(p)), G_i^{\leftarrow}(p) = \inf\{u: G_i(u) \geq p\}.$$

Thus, Lookback-PIT is equivalent to directly selecting the smallest nominal probability level whose empirical lookback mass reaches $p$. For an equal-tailed interval with nominal coverage alpha, the rule is applied at

$$p^- = \frac{1-\alpha}{2}, p^+ = 1 - \frac{1-\alpha}{2}.$$

Both tails therefore use the smallest admissible nominal level under the standard left generalized inverse. This is not an additional heuristic, but follows from the conventional empirical-quantile definition. The equivalence is exact at the transformed cumulative-mass level.

The equivalence is exact at the transformed cumulative-mass level. Numerically, we linearly interpolate the two retrieved futures bracketing the target probability rather than always returning the upper support value. This interpretation also explains the geometry of the adjustment: central lookback ranks tend to contract the interval, two-tailed ranks tend to expand it, and one-sided ranks can translate it.

**F.2. Boundary-space analysis of expansion and shrinkage.** Lookback ranks need not transfer perfectly to future ranks. We therefore temper outward and inward PIT movements separately. Let $r$ index a validation prediction coordinate, $\sigma_s$ be an orientation sign, and $s \in \{-,+\}$ denote the lower and upper tails, with $\sigma_s = -1$ for the lower tail and $\sigma_s = 1$ for the upper tail. Let $B_{r,s}^{\mathrm{B}}, B_{r,s}^{\mathrm{PIT}}, and\ B_{r,s}^{*}$ denote the base, Lookback-PIT, and oracle target boundaries. The oracle boundary is used only for analysis and attains the desired conditional tail probability. Define outward-oriented PIT and oracle displacements by

$$x_{r,s} = \sigma_s\left(B_{r,s}^{\mathrm{PIT}} - B_{r,s}^{\mathrm{B}}\right),$$
$$y_{r,s} = \sigma_s\left(B_{r,s}^{*} - B_{r,s}^{\mathrm{B}}\right).$$

Under this convention, a positive displacement is outward and a negative displacement is inward for either tail. For nonnegative expansion and shrinkage rates b and c, define

$$g_{b,c}(x) = bx(x \geq 0), g_{b,c}(x) = cx(x < 0).$$

The tempered boundary is

$$B_{r,s}^{b,c} = B_{r,s}^{B} + \sigma_s g_{b,c}\left(x_{r,s}\right),$$

and its absolute oracle boundary residual is exactly

$$\left|B_{r,s}^{b,c} - B_{r,s}^{*}\right| = \left|g_{b,c}\left(x_{r,s}\right) - y_{r,s}\right|.$$

Pooling both tails and all validation coordinates gives

$$R_2(b,c) = \sum_r \sum_{s\in\{-,+\}} \left|g_{b,c}\left(x_{r,s}\right) - y_{r,s}\right|$$

A single-rate transformation instead has

$$R_1(a) = \sum_r \sum_{s\in\{-,+\}} \left|ax_{r,s} - y_{r,s}\right|$$

Because the two-rate family contains every single-rate transformation as *b=c=a*,

$$inf_{b,c\geq 0}\{R_2(b,c)\} \leq inf_{a\geq 0}\{R_1(a)\} \leq R_1(1),$$

where *a=1* is untempered Lookback-PIT. Separate rates can therefore attain a boundary residual no larger than the best shared rate or pure PIT. To characterize strict improvement, split the residual into outward and inward components,

$$R_{out}(a) = \sum_{x_{r,s}\geq 0} \left|ax_{r,s} - y_{r,s}\right|$$
$$R_{in}(a) = \sum_{x_{r,s}<0} \left|ax_{r,s} - y_{r,s}\right|$$

By construction, the two-rate objective separates into an outward component and an inward component: $R_2(b,c) = R_{out}(b) + R_{in}(c)$. In contrast, the shared-rate objective requires both components to use the same rate: $R_1(a) = R_{out}(a) + R_{in}(a)$. If the two directional objectives have no common minimizer and don't equal 1, i.e.,

$$arg\min_{a\geq 0} R_{out}(a) \cap arg\min_{a\geq 0} R_{in}(a) = \emptyset,$$
$$arg\min_{a\geq 0} R_1(a) \neq 1$$

then

$$inf_{b,c\geq 0}\{R_2(b,c)\} < inf_{a\geq 0}\{R_1(a)\} < inf\{R_1(1)\}.$$

Hence, directional tempering can strictly reduce the oracle boundary residual whenever outward and inward lookback signals have different transfer reliability. Sharing $b$ and $c$ across the upper and lower tails is a structural regularizer: after orientation, both tails contribute evidence about expansion and contraction, which may reduce estimation variance relative to four tail-specific parameters.

**Connection to quantile and coverage residuals.** Boundary and nominal-level residuals are not numerically identical, but they are locally coupled. Suppose the PIT and oracle boundaries can be represented $B_{r,s}^{\mathrm{B}} = Q_r^{\mathrm{B}}(p_s)$, $B_{r,s}^{\mathrm{PIT}} = Q_r^{\mathrm{B}}\left(u_{r,s}^{X}\right)$, $B_{r,s}^{*} = Q_r^{\mathrm{B}}\left(u_{r,s}^{Y}\right)$, where $u_{r,s}^{X}$ is the lookback-implied probability level and $u_{r,s}^{Y}$ is its oracle future counterpart. The oracle level $u_{r,s}^{Y}$ is introduced only for analysis and is never accessed by the forecasting algorithm. If $Q_r^{\mathrm{B}}$ is locally differentiable, the mean-value theorem gives

$$x_{r,s} = \sigma_s(B_{r,s}^{\mathrm{PIT}} - B_{r,s}^{\mathrm{B}}) = \sigma_s Q_r^{\mathrm{B}\prime}\left(\xi_{r,s}^{X}\right)\left(u_{r,s}^{X} - p_s\right),$$
$$y_{r,s} = \sigma_s(B_{r,s}^{*} - B_{r,s}^{\mathrm{B}}) = \sigma_s Q_r^{\mathrm{B}\prime}\left(\xi_{r,s}^{Y}\right)\left(u_{r,s}^{Y} - p_s\right),$$

for intermediate levels $\xi_{r,s}^{X}$ and $\xi_{r,s}^{Y}$. Boundary displacement is therefore a locally inverse-density-scaled quantile-level displacement: the two have the same direction, while value-space analysis also accounts for the local scale of the predictive distribution.

Let $F_r^{Y}$ be the conditional future CDF with $F_r^{Y}\left(B_{r,s}^{*}\right) = p_s$. Suppose its density is locally bounded by $m_{r,s} \leq f_r^{Y}(z) \leq M_{r,s}$ between $B_{r,s}^{b,c}$ and $B_{r,s}^{*}$. Since $\left|B_{r,s}^{b,c} - B_{r,s}^{*}\right| = \left|g_{b,c}\left(x_{r,s}\right) - y_{r,s}\right|$, the mean-value theorem gives

$$0 < m_{r,s}\left|g_{b,c}\left(x_{r,s}\right) - y_{r,s}\right| \leq \left|F_r^{Y}\left(B_{r,s}^{b,c}\right) - p_s\right|$$
$$\leq M_{r,s}\left|g_{b,c}\left(x_{r,s}\right) - y_{r,s}\right| < +\infty.$$

Thus, under local regularity, boundary residual and tail-probability residual are equivalent up to density-dependent constants. In particular, reducing the boundary residual controls the corresponding tail-probability error; summing the two tail errors yields a bound-on interval-coverage error. This links the boundary-space result to the empirical calibration objective used in our method.

Because oracle boundaries and conditional densities are unavailable in practice, we select b and c directly on validation data by minimizing

$$J_{val}(b,c) = |\hat{e}^{-}_{\mathrm{val}}(b,c) - \gamma| + |\hat{e}^{+}_{\mathrm{val}}(b,c) - \gamma|.$$

where $\gamma = \frac{1-\alpha}{2}$ and the two empirical terms are the lower- and upper-tail miss rates under (*b, c*). Mean interval width breaks ties. The boundary-residual analysis is therefore an oracle explanation for why directional rates can help, whereas the implementation directly optimizes their observable coverage consequence.

The unrestricted analysis requires only nonnegative rates. In the reported implementation, $b \in \{1,1.25,1.5,2,3\}$ and $c \in \{0,0.25,0.5,0.75,1\}$ are reliability-oriented regularizers rather than theoretical requirements: outward PIT corrections are not attenuated, inward corrections are not amplified, val-test drift is relieved by grid search hyperparameters and the validation search space is reduced. The optional expansion cap is likewise a robustness safeguard against isolated extreme boundary movements and is not required by the inequalities above.